\documentclass[11pt,letterpaper]{article}
\usepackage{emnlp2016}
\usepackage{times}
\usepackage{latexsym}
\usepackage{times}
\usepackage{latexsym}
\usepackage{amssymb}
\usepackage{amsthm}
\usepackage[table]{xcolor}
\usepackage{amsmath}
\usepackage{graphicx}
\usepackage{scalerel}
\usepackage{breqn}
\usepackage{url}
\usepackage[rightcaption]{sidecap}
\usepackage{enumitem}
\setitemize{noitemsep,topsep=10pt,parsep=0pt,partopsep=0pt}
\setenumerate{noitemsep,topsep=10pt,parsep=0pt,partopsep=0pt}

\emnlpfinalcopy

\def\emnlppaperid{***}

\newcommand{\ns}[1]{\textcolor{blue}{\bf\small [#1 --NS]}}
\newcommand{\nascomment}[1]{\textcolor{blue}{\bf\small [#1 --NS]}}
\newcommand{\cjd}[1]{\textcolor{pink}{\bf\small [#1 --CJD]}}

\newcommand{\kk}[1]{\textcolor{orange}{\bf\small [#1 --KK]}}

\newcommand{\ignore}[1]{}

\newcommand{\Lab}{$\mathit{Lab}$}
\usepackage{enumitem}

\usepackage{enumitem}
\setitemize{noitemsep,topsep=6pt,parsep=0pt,partopsep=0pt,itemindent=0pt,leftmargin=12pt,labelwidth=12pt}
\setenumerate{noitemsep,topsep=6pt,parsep=0pt,partopsep=0pt,itemindent=0pt,leftmargin=12pt,labelwidth=12pt}

\title{Character Sequence Models for Colorful Words}

\author{Kazuya Kawakami~$^{\spadesuit}$, Chris Dyer$^{\spadesuit\clubsuit}$ ~ Bryan R.~Routledge$^{\diamondsuit}$ ~ Noah A. Smith$^{\heartsuit}$\\
$^{\spadesuit}$School of Computer Science, Carnegie Mellon University,
Pittsburgh, PA, USA \\
$^{\clubsuit}$Google DeepMind, London, UK \\
$^{\diamondsuit}$Tepper School of Business, Carnegie Mellon University, Pittsburgh, PA, USA \\
$^{\heartsuit}$Computer Science \& Engineering, University of Washington, Seattle, WA, USA\\
{\small \tt \{kkawakam,cdyer\}@cs.cmu.edu, routledge@cmu.edu, nasmith@cs.washington.edu}
}

\date{}

\begin{document}

\maketitle

\ignore{
\ns{note new title; I think ``compositional'' is confusing/misleading} \kk{Yes.}
}

\begin{abstract}
We present a neural network architecture to predict a point in 
color space from the sequence of characters in the color's name.  Using large scale color--name
pairs obtained from an online color design forum, we evaluate our
model on a ``color Turing test'' and find that, given a name,
 the colors predicted by our model are preferred by annotators to color names created by humans. 
Our datasets and demo system are available online at
\url{http://colorlab.us}.
\end{abstract}

\section{Introduction}
Color is a valuable vehicle for studying the association between words and their nonlinguistic referents. Perception of color has long been studied in psychology, and quantitative models linking physical stimuli and psychological perception have been in place since the 1920s \cite{broadbent:2004}. Although perceptually faithful color representations require only a few dimensions (\S\ref{sec:colorspace}), linguistic expressions of color often rely on association and figurative language.  
There are, for example, 34,000 examples of ``blue'' in our data.  The
varieties of blue range can be emotional,
descriptive, metaphoric, literal, and whimsical.  \ignore{As examples,} 
Consider these examples (best viewed in color):
\textcolor[rgb]{0.34765625,  0.37109375,  0.5390625}{murkey blue}, 
\textcolor[rgb]{0.6015625 ,  0.76953125,  0.92578125}{blueberry muffin}, 
\textcolor[rgb]{0.4453125 ,  0.81640625,  0.65234375}{greeny blue}, and
\textcolor[rgb]{0.375     ,  0.1796875 ,  0.86328125}{jazzy blue}.

%

This rich variety of descriptive names of colors provides an ideal way to
study linguistic creativity, its variation, and an important aspect
of visual understanding.  This paper uses predictive modeling to
explore the relationship between colors (represented in three dimensions) and casual, voluntary linguistic
descriptions of them by users of a crafting and design website
(\S\ref{sec:dataset}).\footnote{http://www.colourlovers.com}

In this dataset's creative vocabulary, word-level representations are
so sparse as to be useless, so we turn to models that build name
representations out of \emph{characters}
(\S\ref{sec:w2c}).
We evaluate our model on a ``color Turing test'' and find that,
given a name, it tends to generate a color that humans find matches the name better than the color that actually inspired the name. We also investigate the reverse mapping, from colors to names
(\S\ref{sec:c2w}). We compare a conditional LSTM language model used
in caption generation \cite{karpathy2014deep} to a new latent-variable
model, achieving a 10\% perplexity reduction.

We expect such modeling to find purchase in computational creativity applications~\cite{vealeunweaving}, design and marketing aids~\cite{deng2010consumer}, and new methods for studying the interface between the human visual and linguistic systems~\cite{marcus1991graphic}.

\section{Color Spaces}\label{sec:colorspace}

In electronic displays and other products, colors are commonly
represented in RGB space where each color is embedded in
$\{0,\ldots,255\}^3$, with coordinates corresponding to red, green,
and blue levels. While convenient for digital processing, distances in this space are perceptually non-uniform. We instead use a different three-dimensional representation, \Lab, which  was
originally designed so that Euclidean distances correlate with
human-perceived differences \cite{hunter1958photoelectric}.  \Lab{} is
also continuous, making it more suitable for the gradient-based
learning used in this paper.  The transformation from RGB to \Lab{} is nonlinear.

\section{Task and Dataset}\label{sec:dataset}
We consider the task of predicting a color in \Lab{} space
given its name. \ignore{(i) predicting a color (in \Lab{} space)
given its name, and (ii) the reverse (which we treat as a conditional language
modeling problem).}  Our dataset is a collection of user-named colors downloaded from
COLOURlovers,\footnotemark[1] a creative community where
people create and share colors, palettes, and
patterns. Our dataset contains 776,364 pairs with 581,483 unique names.
Examples of the color/name pairs from COLOURlovers are the following:
\textcolor[rgb]{0.9960, 0.2627, 0.3961}{Sugar Hearts You},
\textcolor[rgb]{1,0.6,0}{Vitamin C},
\textcolor[rgb]{0.8039 , 0.8431, 0.7137}{Haunted milk}.

We considered two held-out datasets from other sources; these do not
overlap with the training data. \\ \textbf{ggplot2}: the 141 officially-named colors used in ggplot2, a common plotting
package for the R programming language (e.g.,
\textcolor[rgb]{0.09803922,  0.09803922,  0.43921569}{MidnightBlue}.
\textcolor[rgb]{0.23529412,  0.70196078,
  0.4431372}{MediumSeaGreen}),\footnote{http://sape.inf.usi.ch/quick-reference/ggplot2/colour}\\ \textbf{Paint}: The paint manufacturer Sherwin Williams has 7,750
named colors (e.g., \textcolor[rgb]{0.58039216,  0.        ,  0.}{Pompeii Red}, \textcolor[rgb]{0.95686275,  0.88235294,  0.63137255}{Butter Up}).\footnote{http://bit.ly/PaintColorNames}

\ignore{
 Table~\ref{tb:dataset} summarizes our data.

\begin{figure}[ht]
\begin{center}
\small
\begin{tabular}{|c|c|c|}\hline
\cellcolor[rgb]{0.3333 , 0.3843, 0.4392} \textcolor{white}{Mighty Slate} &
\cellcolor[rgb]{0.8039 , 0.8431, 0.7137}{Haunted milk} &
\cellcolor[rgb]{0.9982 , 0.9843, 0.8902}{vanilla cream} \\ \hline
\cellcolor[rgb]{0.7647, 1, 0.4078}{certain frogs} &
\cellcolor[rgb]{1,0,0}{Red} &
\cellcolor[rgb]{0.2, 0.2, 0.2} \textcolor{white}{Grr-ey} \\ \hline
\cellcolor[rgb]{1, 0.6, 0}{Vitamin C} &
\cellcolor[rgb]{0.8275,0.0980 ,0.5882}{Bloons} &
\cellcolor[rgb]{0.9960, 0.2627, 0.3961}{Sugar Hearts You} \\ \hline
\end{tabular}
\end{center}
\caption{Examples of color/name pairs from
  COLOURlovers. \ignore{\nascomment{names were too small to read, so I redid
    this in latex}}
\label{fig:colorlover}}
\end{figure}
}

\begin{table}[t]
\begin{center}\small
\begin{tabular}{l|rr}
\hline
       & Number of pairs & Unique names\\\hline
Train        & 670,032         & 476,713     \\
Dev.         & 53,166          & 52,753      \\
Test         & 53,166          & 52,760      \\\hline
ggplot2      & 66              & 66          \\
Paint        & 956             & 956         \\
\ignore{Brand        & 814     & 813         \\}\hline
\end{tabular}
\end{center}
\caption{Datasets used in this paper.  The train/dev./test split of
  the COLOURlovers data was random. For ggplot2 and  Paint, we show the number of test instances which are not in Train set.
\label{tb:dataset}}
\end{table}

\section{Names to Colors}\label{sec:w2c}
Our word-to-color model is used to predict a color in
\Lab{} space given the sequence of characters in a color's name, $\boldsymbol{c} = \langle
c_1, c_2, \ldots, c_{|\boldsymbol{c}|}\rangle$, where each $c_i$ is a
character in a finite alphabet.  Each character $c_i$ is represented
by learned vector embedding in $\mathbb{R}^{300}$. To build a color
out of the sequence, we use an LSTM \cite{hochreiter1997long} with
300 hidden units. The final hidden state is used as a vector representation $\mathbf{h} \in \mathbb{R}^{300}$ of the sequence. The associated color value in \Lab{} space is then defined to be
$\hat{\mathbf{y}} = \sigma(\mathbf{W}\mathbf{h} + \mathbf{b})$,
where $\mathbf{W} \in \mathbb{R}^{3 \times 300}$ and $\mathbf{b} \in \mathbb{R}^3$ transform $\mathbf{h}$.

This model instantiates the one proposed by
\newcite{ling2015character} for learning word embeddings built from
representations of characters.\ignore{\footnote{This model makes the somewhat counterintuitive assumption that the letters that a word consists of can be composed together to learn word representations. However, the nonlinear dynamics of the LSTM are quite effective at learning the idiosyncratic compositions of characters necessary to model differences between words.}}

To learn the parameters of the model (i.e., the parameters of the LSTMs, the character embeddings, and $\mathbf{W}$ and $\mathbf{b}$), we use reference color labels $\mathbf{y}$ from our training set and minimize squared error,
$|| \mathbf{y} - \hat{\mathbf{y}}||^2$, averaged across the training
set.  Learning is accomplished using backpropagation and the Adam update rule \cite{kingma2014adam}.

\subsection{Evaluation}
We evaluated our model in two ways. First, we computed mean-squared
error on held-out data using several variants of our model. The
baseline models are linear regression models,  which
predict a color from a bag of character unigrams and bigrams. We
compare an RNN and LSTMs with one and two layers. Table~\ref{tb:regression} shows that the two-layer LSTM achieves lower error than the
unigram and bigram baselines and an RNN. We see the same pattern of results on the out-of-domain test sets.

\begin{table}[t]
\begin{center}\small
\begin{tabular}{l|rrrr}
\hline
Model       & Test    & ggplot2    & Paint  \ignore{& Brand}    \\\hline
Unigram     & 1018.35 & 814.58     & 351.54 \ignore{& 1230.91}\\
Bigram      & 977.46  & 723.61     & 364.41 \ignore{& 1233.10}\\
RNN         & 750.26  & 431.90     & 305.05 \ignore{& 1280.06}\\
1-layer LSTM        & 664.11  & 355.56     & 303.03 \ignore{& 1259.99}\\
2-layer LSTM   & 652.49  & 343.97     & 274.83 \ignore{& 1242.82}\\\hline
\end{tabular}
\end{center}
\caption{MSE in \Lab{} space on held-out datasets. \label{tb:regression}}
\end{table}

\paragraph{The Color Turing Test.} Our second evaluation attempts to
assess whether our model's associations are human-like. For this
evaluation, we asked human judges to choose the color better described
by a name from one of our test sets:  our model's predicted color or
the color in the data. For each dataset, we randomly selected 20
examples. \ignore{\kk{For test set, we selected 20 samples where
    \Lab{} distance between original color and our predictions are
    large since randomly selected samples included instances where the
    generated color is very close to actual color.}\ignore{\ns{and
      what about our test set?}}} 111 judges considered each
instance.\footnote{We excluded results from an additional 19
  annotators who made more than one mistake in a color blindness
  test~\cite{oliver1888tests}.}  Judges were presented
instances in random order and forced to make a choice between the
two and explicitly directed to make an arbitrary choice if neither was
better.\footnote{A preliminary study that allowed a judge to say that
  there was no difference led to a similar result.} The test is shown at
\url{http://colorlab.us/turk}.  

\ignore{\ns{NAACL R2:   (Obvious?) Question: are these 20 examples cases
for which the generated color is different than the actual one? }\kk{Yes. Thats' why we didn't do preprocessing for brand and paint.}}

Results are shown in Table~\ref{tb:turing}; on the ggplot2 and Paint
datasets, our prediction is preferred to the actual names in a
majority of cases. The Test dataset from COLOURlovers is a little bit challenging, with more noisy and creative names; still, in the majority of cases, our prediction is preferred.

\begin{table}[t]
\begin{center}
\begin{tabular}{l|rrr}
\hline
Preference          & Test     & ggplot2 & Paint\\\hline
Actual color        & 43.2\%   & 32.6\%  & 31.0\%\\
Predicted color     & 56.7\%   & 67.3\%  & 69.0\%\\\hline
\end{tabular}
\end{center}
\caption{Summary of color Turing test results. \label{tb:turing}}
\end{table}

\subsection{Visualization and Exploration}

To better understand our model, we provide illustrations of its predictions on several kinds of inputs.

\begin{figure}[ht]
\begin{center}
\includegraphics[width=0.9\linewidth]{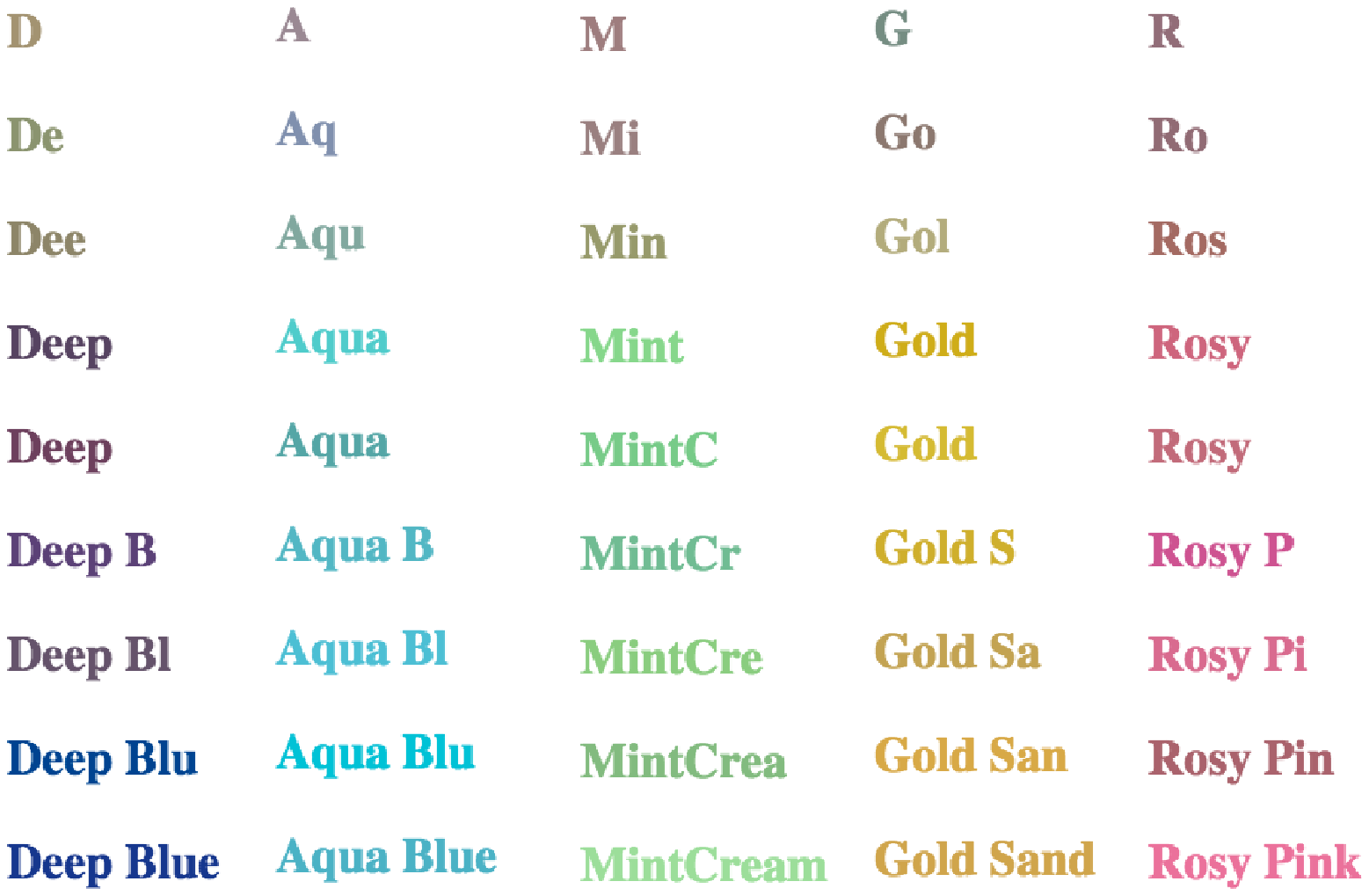}
\end{center}
\vspace{-.4cm}\caption{Visualization of character-by-character
  prediction. \ignore{\nascomment{too small.  can we render in latex instead?}}
\label{fig:example}}
\end{figure}

\paragraph{Character by character prediction.}
We consider how our model reads color names character by
character. Fig.~\ref{fig:example} shows some examples, such as
\emph{blue}, variously modified.
The word \textit{deep} starts dark brown, but eventually modifies
\emph{blue} to a dark blue.  Our model also performs sensibly on colors
named after things (\emph{mint}, \emph{cream}, \emph{sand}).

\begin{figure}
\includegraphics[width=1\linewidth]{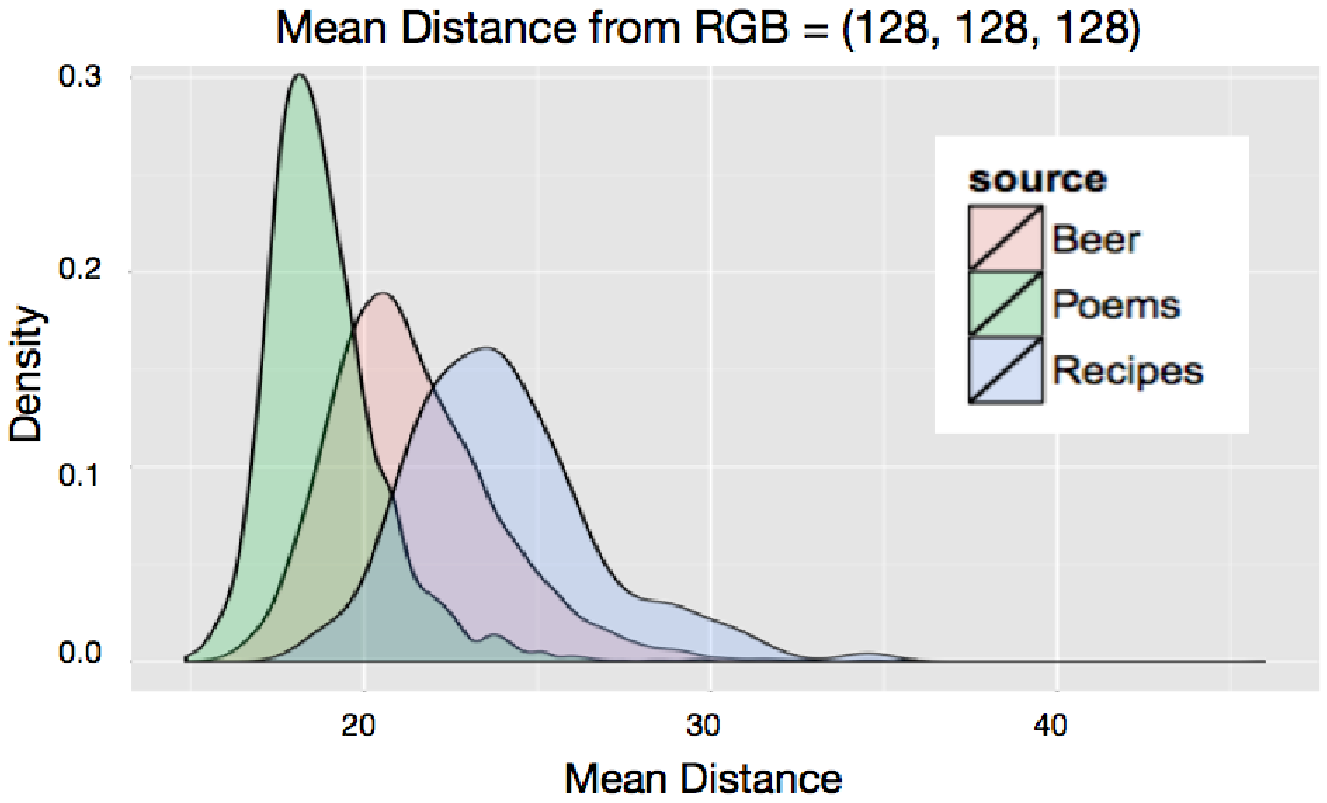}
\caption{Distribution of Euclidean distances in \Lab{} from estimated colors of
  words in each corpus to RGB (128, 128, 128).
\label{fig:poem}}
\end{figure}

\begin{figure}[ht]
\vspace{-.4cm}
\begin{center}
\includegraphics[width=1\linewidth]{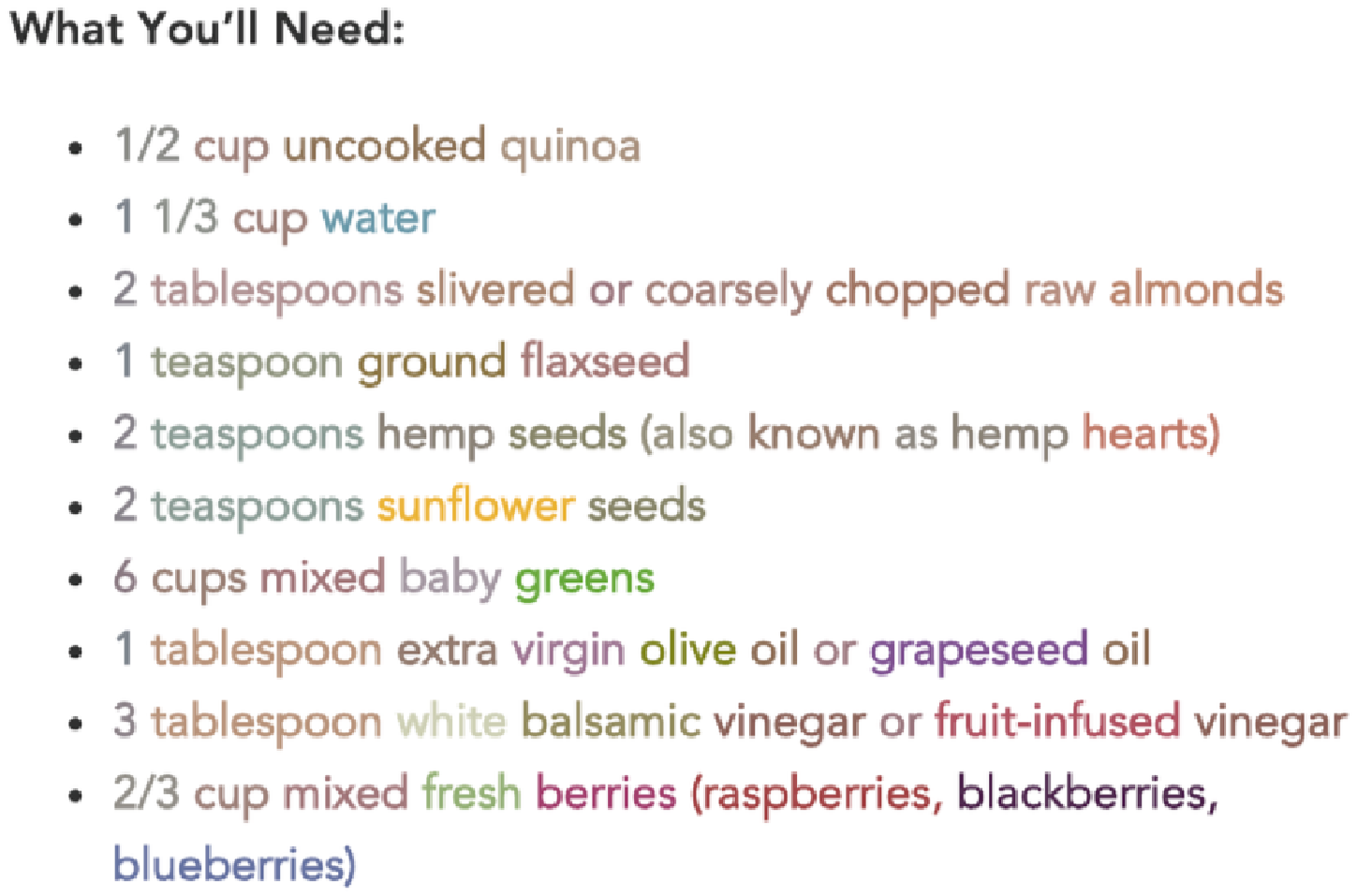}
\end{center}
\vspace{-.6cm}\caption{A recipe from
\texttt{greatist.com}.
\label{fig:recipe}}
\end{figure}

\paragraph{Genre and color.}
We can use our model to investigate how colors are evoked in text by
predicting the colors of each word in a text. Fig.~\ref{fig:recipe}
shows a colored
recipe. 
Noting that many words are rendered in neutral grays and tans, we
investigated how our model colors words in three corpora:  3,300
English poems (1800--present), 256 recipes from the CURD
dataset \cite{tasse2008sour},\footnote{http://www.cs.cmu.edu/~ark/CURD/} and 6,000
beer reviews.\footnote{http://beeradvocate.com}  For each
corpus,
we examine the distribution of Euclidean distances of $\mathbf{\hat{y}}$ from
the \emph{Lab} representation of the ``middle'' color RGB (128, 128, 128). The Euclidean distances from the mean are measuring the variance of the color of words in a document. Fig.~\ref{fig:poem} shows these distributions; recipes and beer
reviews are more ``colorful'' than poems, under our model's learned
definition of color.

\section{Generating Names from Colors}\label{sec:c2w}
The first of our two color naming models generates character sequences conditioned on \Lab{} color
representations, following other sequence-to-sequence approaches
\cite{sutskever2014sequence,karpathy2014deep}.  The transformation
is as follows:  First, a linear transformation maps the color vector
into 300 dimensions, together comprising the initial hidden and memory vectors.  Next a character LSTM is iteratively applied to the hidden, memory, and next-character vectors, and the next character produced by applying affine and then softmax functions to the hidden vector. The model is trained to maximize conditional likelihood of each character given its history. We used 300 dimensions for character embeddings and recurrence weights. The output vocabulary size was 98 without lowercasing.

\ignore{\begin{figure*}[ht]
\begin{center}
\includegraphics[width=0.5\linewidth]{img/generation.eps}
\end{center}
\caption{Description of the operation to modulate sense from ambiguous type level vector.}
\label{fig:generation}
\end{figure*}}

We also propose a model to capture variations in color description with latent
variables by extending the variational autoencoder
\cite{kingma2013auto} to a conditional model. We want to model the
conditional probability of word $\mathbf{y}$ and latent variables
$\mathbf{z}$ \ignore{\nascomment{what are the latent variables?  I don't
  understand this model at all, and I'm sure reviewers won't either.}} given
color $\mathbf{x}$. The latent variable gives the model capacity to account for the complexity of the color--word mapping. Since $p(\mathbf{y},\mathbf{z}\mid\mathbf{x}) =
p(\mathbf{z})p(\mathbf{y}\mid\mathbf{x},\mathbf{z})$, the variational
objective is: \ignore{
{\small  
\begin{align*}
& \mathbb{E}_{q_{\phi}(\mathbf{z}\mid \mathbf{x})}[-\log q_{\phi}(\mathbf{z}\mid \mathbf{x}) + \log p_{\theta}(\mathbf{y},\mathbf{z}\mid \mathbf{x})]\\
&= \mathbb{E}_{q_{\phi}(\mathbf{z}\mid \mathbf{x})}[-\log q_{\phi}(\mathbf{z}\mid \mathbf{x}) + \log p_{\theta}(\mathbf{y}\mid \mathbf{x}, \mathbf{z})p_{\theta}(\mathbf{z})]\\
&\simeq -D_{KL}(q_\phi(\mathbf{z}\mid \mathbf{x}) \mid \mid  p_{\theta}(\mathbf{z})) + \frac{1}{L}\sum_{l=1}^{L} \log p_{\theta}(\mathbf{y}\mid \mathbf{x}, \mathbf{z}^{l})
\end{align*}
}
}
{\small  
\begin{align*}
& \mathbb{E}_{q_{\phi}(\mathbf{z}\mid \mathbf{x})}[-\log q_{\phi}(\mathbf{z}\mid \mathbf{x}) + \log p_{\theta}(\mathbf{y},\mathbf{z}\mid \mathbf{x})]\\
&= \mathbb{E}_{q_{\phi}(\mathbf{z}\mid \mathbf{x})}[-\log q_{\phi}(\mathbf{z}\mid \mathbf{x}) + \log p_{\theta}(\mathbf{y}\mid \mathbf{x}, \mathbf{z})p(\mathbf{z})]\\
&\simeq -D_{KL}(q_\phi(\mathbf{z}\mid \mathbf{x}) \mid \mid  p(\mathbf{z})) + \frac{1}{L}\sum_{l=1}^{L} \log p_{\theta}(\mathbf{y}\mid \mathbf{x}, \mathbf{z}^{l})
\end{align*}
}
The first term regularizes the shape of posterior, $q(\mathbf{z}\mid
\mathbf{x})$, to be close to prior $p(\mathbf{z})$ where it is a
Gaussian distribution, $p(\mathbf{z})= \mathcal{N}(\mathbf{0},
\mathbf{I})$. The second term is the  log likelihood of  the character
sequence conditioned on color values. To optimize $\theta$ and $\phi$,
we reparameterize the model, we write $\mathbf{z}$ in terms of a mean
and variance and samples from a standard normal distribution, i.e.,
$\mathbf{z} = \mu + \sigma\epsilon$ with $\epsilon \sim
\mathcal{N}(\mathbf{0},\mathbf{I})$. We predict mean and log variance of the model with a multi-layer perceptron
and initialize the decoder-LSTM with $\mathbf{h}_{0} = \tanh(\mathbf{Wz} + \mathbf{b})$. We trained the model with mini-batch size 128 and Adam optimizer. The sample size $L$ was set to 1. \ignore{\cjd{what is $p_{}$}\kk{It's gaussian prior.}}

\ignore{\begin{eqnarray*}
\log p(\mathbf{y}|\mathbf{x})
&=& \int q(\mathbf{z}|\mathbf{x}) \log \frac{q(\mathbf{z}|\mathbf{x})}{p(\mathbf{z}|\mathbf{x},\mathbf{y})} d\mathbf{z} + \int q(\mathbf{z}|\mathbf{x}) \log \frac{p(\mathbf{y}, \mathbf{z}|\mathbf{x})}{q(\mathbf{z}|\mathbf{x})} d\mathbf{z}\\
&=& D_{KL}(q(\mathbf{z}|\mathbf{x}) || p(\mathbf{z}|\mathbf{x},\mathbf{y})) + \mathcal{L}(\theta, \phi; \mathbf{x},\mathbf{y})
\end{eqnarray*}

\begin{eqnarray*}
\mathcal{L}^{\alpha}(\theta, \phi; x,y) &=& \mathbb{E}_{q_{\phi}(z|x)}[-\log q_{\phi}(z|x) + \log p_{\theta}(y,z|x)]\\
&=& \mathbb{E}_{q_{\phi}(z|x)}[-\log q_{\phi}(z|x) + \log p_{\theta}(y|x,z)p_{\theta}(z)]\\
&\simeq& \frac{1}{L} \sum_{l=1}^{L} \log p(y,z^{l}|x) - \log q_{\phi}(z^{l})
\end{eqnarray*}

\begin{eqnarray*}
\mathcal{L}^{\beta}(\theta, \phi; x,y) &=& \mathbb{E}_{q_{\phi}(z|x)}[-\log q_{\phi}(z|x) + \log p_{\theta}(y,z|x)]\\
&=& \mathbb{E}_{q_{\phi}(z|x)}[-\log q_{\phi}(z|x) + \log p_{\theta}(y|x, z)p_{\theta}(z)]\\
&\simeq& -D_{KL}(q(z|x) || p_{\theta}(z)) + \frac{1}{L}\sum_{l=1}^{L} \log p_{\theta}(y|x, z^{l})
\end{eqnarray*}}

\paragraph{Evaluation.} We evaluated our models by estimating
perplexity on the Test set (Table~\ref{tb:dataset})\ignore{\ns{include all test sets?}\kk{no only colorlover}}. Our baseline is a character-level unconditional LSTM language model. Conditioning on color improved per-character perplexity by 7\% and the  latent variable gave a further 3\%; see Table~\ref{tb:lm}.

A second dataset we evaluate on is the Munroe Color
Corpus~\cite{munroe} which contains 2,176,417 color description for 829 words (i.e., single
words have multiple color descriptions).
\newcite{monroe2016learning} have developed word-based (rather character-based) recurrent neural network model.

 Our character-based model
with 1024 hidden units achieved 12.48 per-description perplexity,
marginally better than 12.58 obtained with a word-based neural network model
reported in that work. Thus, we see that modeling color names as sequences of characters is wholly feasible. However, since the corpus only contains color
description for 829 words, the model trained on the Munroe Color Corpus does not provide suitable supervision for evaluation on our more lexically diverse dataset.

\begin{table}[t]
\begin{center}
\begin{tabular}{l|r}
\hline
Model          & Perplexity \\\hline
LSTM-LM        & 5.9        \\
VAE            & 5.9        \\\hline
color-conditioned LSTM-LM       & 5.5        \\
color-conditioned  VAE           & \bf{5.3}   \\\hline
\end{tabular}
\end{center}
\caption{Comparison of language models.
\label{tb:lm}}
\end{table}

\ignore{
\begin{table}[t]
\begin{center}
\begin{tabular}{l|r}
\hline
Model                       & Perplexity\\\hline
HM                          & 14.41\\
LUX                         & 13.61\\
Word LSTM                   & 12.58\\
Char LSTM                   & 12.41\\
Char VAE                    & 12.30\\\hline
\end{tabular}
\end{center}
\caption{Comparison of language models.
\label{tb:lm}}
\end{table}
}

\section{Related Work and Discussion}

Color is one of the lowest-level visual signals playing an important
role in cognition~\cite{wurm1993color} and
behavior~\cite{maier2008mediation,lichtenfeld2009semantic}.  It plays
a role in human object recognition: to name an object, we first need to encode
visual information such as shape and surface
information including color and texture. Given a visual encoding, we
search our memory for a structural, semantic and phonological
description~\cite{humphreys1999objects}. Adding color information to shape significantly improves naming
 accuracy and speeds 
correct response times~\cite{rossion2004revisiting}. 

Colors and their names have some association in our
cognition. The Stroop \shortcite{stroop1935studies} effect is
a well-known example showing interference of colors and color terms:
when we see a color term printed in a different color---\textcolor{red}{blue}---it takes us longer to name the word, and we
are more prone to naming errors than when the ink matches---\textcolor{blue}{blue}
\cite{de2003role}.

Recent evidence suggests that colors and words are associated in the
brain. The brain uses different regions to perceive
various modalities, but processing a
color word activates the same brain region as the color it denotes~\cite{del2006category,simmons2007common}.

Closer to NLP, the relationship between visual stimuli and their linguistic
descriptions by humans has been explored extensively through automatic
text generation from images
\cite{kiros2014multimodal,karpathy2014deep,xu2015show}. Color
association with word semantics has also been investigated in several previous papers
\cite{mohammad2011colourful,heer:2012,andreas2014grounding,mcmahan2015bayesian}.


\section{Conclusion}
In this paper, we introduced a computational model to predict a point in 
color space from the sequence of characters in the color's name.
Using a large set of color--name
pairs obtained from a color design forum, we evaluate our model on a ``color Turing test'' and find that, given a name,
 the colors predicted by our model are preferred by annotators to color names created by humans. We also investigate the reverse mapping, from colors to names. We compare a conditional LSTM language model to a new latent-variable model, achieving a 10\% perplexity reduction.

\section*{Acknowledgments}
We thank Lucas Beyer for very helpful comments and discussions, and we also appreciate all the participants of our color Turing test.

\bibliography{emnlp2016}
\bibliographystyle{emnlp2016}

\end{document}